\documentclass[lettersize,journal]{IEEEtran}
\usepackage{amsmath,amsfonts}
\usepackage{algorithm}
\usepackage{array}

\usepackage{textcomp}
\usepackage{stfloats}
\usepackage{url}
\usepackage{comment}
\usepackage{graphicx}
\usepackage{tabularx}
\usepackage{amsmath,amssymb,amsfonts, bm}
\usepackage{multicol}
\usepackage{colortbl}
\usepackage[most]{tcolorbox}
\usepackage{algpseudocode}
\usepackage{graphics}
\usepackage{verbatim}
\usepackage[font=small]{caption}
\usepackage{subcaption}
\usepackage{cite}
\usepackage{multirow}
\usepackage[xindy,nonumberlist,nogroupskip]{glossaries}
\makeglossaries
\newacronym{FL}{FL}{Federated Learning}
\newacronym{DLG}{DLG}{Deep Leakage from Gradients}
\newacronym{SecAgg}{SecAgg}{Secure Aggregation}
\newacronym{FedAvg}{FedAvg}{Federated Average}
\newacronym{MPC}{MPC}{Multi-Party Computation}
\newacronym{DMS}{DMS}{Distributed Markovian Switching}
\newacronym{MSE}{MSE}{Mean Squared Error}
\newacronym{FL-BNN}{FL-BNN}{Federated Learning-based Bayesian Neural Networks}
\newacronym{STLF}{STLF}{Short-term Load Forecasting}
\newacronym{LTC}{LTC}{Learning-Then-Consensus}
\newacronym{CTL}{CTL}{Consensus-Then-Learning}
\newacronym{P2P}{P2P}{Peer-to-Peer}
\newacronym{DFC}{DFC}{Distributed Learning with Fully-Connected}
\newacronym{DRING}{DRING}{Distributed Learning with Ring Topology}
\newacronym{DP}{DP}{Differential Privacy}

\newcommand{\Mat}[1]{\textbf{#1}}

\newtheorem{assumption}{\bf Assumption}

\newtheorem{theorem}{\bf Theorem}
\newtheorem{remark}{\bf Remark}

\newenvironment{proof}{{\bf {Proof:}}}{\hfill $\blacksquare$} 
\usepackage{color}
\newcommand{\share}[1]{\langle #1 \rangle}
\newcount\Comments  
\usepackage{xcolor}
\definecolor{darkgreen}{rgb}{0,0.5,0}
\definecolor{purple}{rgb}{1,0,1}
\newcommand{\kibitz}[2]{\ifnum\Comments=1\textcolor{#1}{#2}\fi}

\newcommand{\yi}[1]  {\kibitz{purple}   {#1}}

\newcommand{\tas}[1]{\kibitz{red} }
\newcommand{\R}[1]{{\color{black}{#1}}}
\newcommand{\RR}[1]{{\color{black}{#1}}}
\usepackage{enumitem}

\begin{document}

\title{Privacy-Preserving Distributed Learning for Residential Short-Term Load Forecasting}

\author{Yi Dong$^*$$^\S$, Yingjie Wang$^*$$^\dag$, Mariana Gama$^\ddag$, Mustafa A. Mustafa$^\mathparagraph$$^\ddag$, Geert Deconinck$^\dag$, and Xiaowei Huang$^\S$
\thanks{* Both authors contributed equally to this research.}
\thanks{$^\S$ Department of Computer Science, University of Liverpool, UK, \{yi.dong, xiaowei.huang\}@liverpool.ac.uk}
\thanks{$^\dag$ Electa, Department of Electrical Engineering (ESAT), KU Leuven and EnergyVille, Belgium, \{tony.wang, geert.deconinck\}@kuleuven.be}
\thanks{$^\ddag$ COSIC, Department of Electrical Engineering (ESAT), KU Leuven, Belgium, mariana.botelhodagama@kuleuven.be}
\thanks{$^\mathparagraph$ Department of Computer Science, The University of Manchester, UK, mustafa.mustafa@manchester.ac.uk}
\thanks{This work is supported by the UK EPSRC (End-to-End Conceptual Guarding of Neural Architectures [EP/T026995/1]) and the FWO SBO project SNIPPET (Secure and Privacy-Friendly Peer-to-Peer Electricity Trading [S007619N]). This project has received funding from the European Union’s Horizon 2020 (FOCETA: grant
agreement No 956123) and UKRI (SPACE: project No 10046257).}
}
\markboth{IEEE INTERNET OF THINGS JOURNAL,~Vol.~XX, No.~X, February~2024}%
{Shell \MakeLowercase{\textit{et al.}}: A Sample Article Using IEEEtran.cls for IEEE Journals}

\IEEEpubid{\begin{minipage}{\textwidth}\ \centering
    Copyright \copyright 2024 IEEE. Personal use of this material is permitted. \\However, permission to use this material for any other purposes must be obtained from the IEEE by sending a request to pubs-permissions@ieee.org.
\end{minipage}}

\maketitle

\begin{abstract}
In the realm of power systems, the increasing involvement of residential users in load forecasting applications has heightened concerns about data privacy. Specifically, the load data can inadvertently reveal the daily routines of residential users, thereby posing a risk to their property security. While federated learning (FL) has been employed to safeguard user privacy by enabling model training without the exchange of raw data, these FL models have shown vulnerabilities to emerging attack techniques, such as Deep Leakage from Gradients and \R{poisoning attacks}. To counteract these, we initially employ a Secure-Aggregation (SecAgg) algorithm that leverages multiparty computation cryptographic techniques to mitigate the risk of gradient leakage. However, the introduction of SecAgg necessitates the deployment of additional sub-center servers for executing the multiparty computation protocol, thereby escalating computational complexity and reducing system robustness, especially in scenarios where one or more sub-centers are unavailable. To address these challenges, we introduce a Markovian Switching-based distributed training framework, the convergence of which is substantiated through rigorous theoretical analysis. \R{The Distributed Markovian Switching (DMS) topology shows strong robustness towards the poisoning attacks as well.} Case studies employing real-world power system load data validate the efficacy of our proposed algorithm. It not only significantly minimizes communication complexity but also maintains accuracy levels comparable to traditional FL methods, thereby enhancing the scalability of our load forecasting algorithm.
\end{abstract}


\section{Introduction}

\IEEEPARstart{E}{lectric} load forecasting plays an essential role in power scheduling, planning, operating and management \cite{abedinia2016new, ali2019optimum}. 
The stability of the power system is under threat due to the intermittence of renewable energy generations and the complex nature of utility-customer interactions and dynamic behaviors. To overcome this, residential \gls{STLF} has been widely studied to facilitate the power system operations \cite{kong2017short,lin2021spatial}. 
However, it is evident that residential-user privacy is at risk when residential-user load data is collected and mined \cite{nightingale2022effect}.
For example, the residential user's daily routines and presence at home can be detected with a high probability from its electric load data, which will directly affect the residential user's property safety.
Therefore, how to accurately forecast residential power load while ensuring data privacy becomes an open challenge. 


\IEEEpubidadjcol
The \gls{FL} method has been introduced in recent years to overcome the challenges of residential-user privacy. It can decouple the data storage from the training process \cite{li2020federated}, while reaching a desirable accuracy compared to the centrally trained model \cite{gholizadeh2022federated,fernandez2022privacy}. There are already efforts to apply \gls{FL} to power system forecasting to preserve the sensitive individual consumption profiles \cite{yang2023integrated, tan2022towards,lin2021privacy, gao2021decentralized,savi2021short,fekri2022distributed}. 
Yang \textit{et al.} integrate variational mode decomposition, federated k-means clustering algorithm, and SecureBoost together for \gls{STLF} with data privacy protection \cite{yang2023integrated}. 
Jun \textit{et al.} propose a novel method for disaggregating community-level behind-the-meters solar generation using a federated learning-based Bayesian neural network, which can preserve the confidentiality of utilities' data \cite{lin2021privacy}.  
Yong \textit{et al.} propose a verifiable and oblivious secure aggregation for \gls{FL} \cite{9969897}. Their algorithm could tolerate the high drop-up rate of clients during large-scale \gls{FL} training. 
Asad \textit{et al.} highlight recent \gls{FL} algorithms and evaluate their communication efficiency in detailed comparisons. The experimental results indicate that training data-driven models using \gls{FL} not only enhances security and privacy, but also reduces communication costs \cite{9437738}.

\begin{figure}[htbp]
	\centering
	\includegraphics[width=\linewidth]{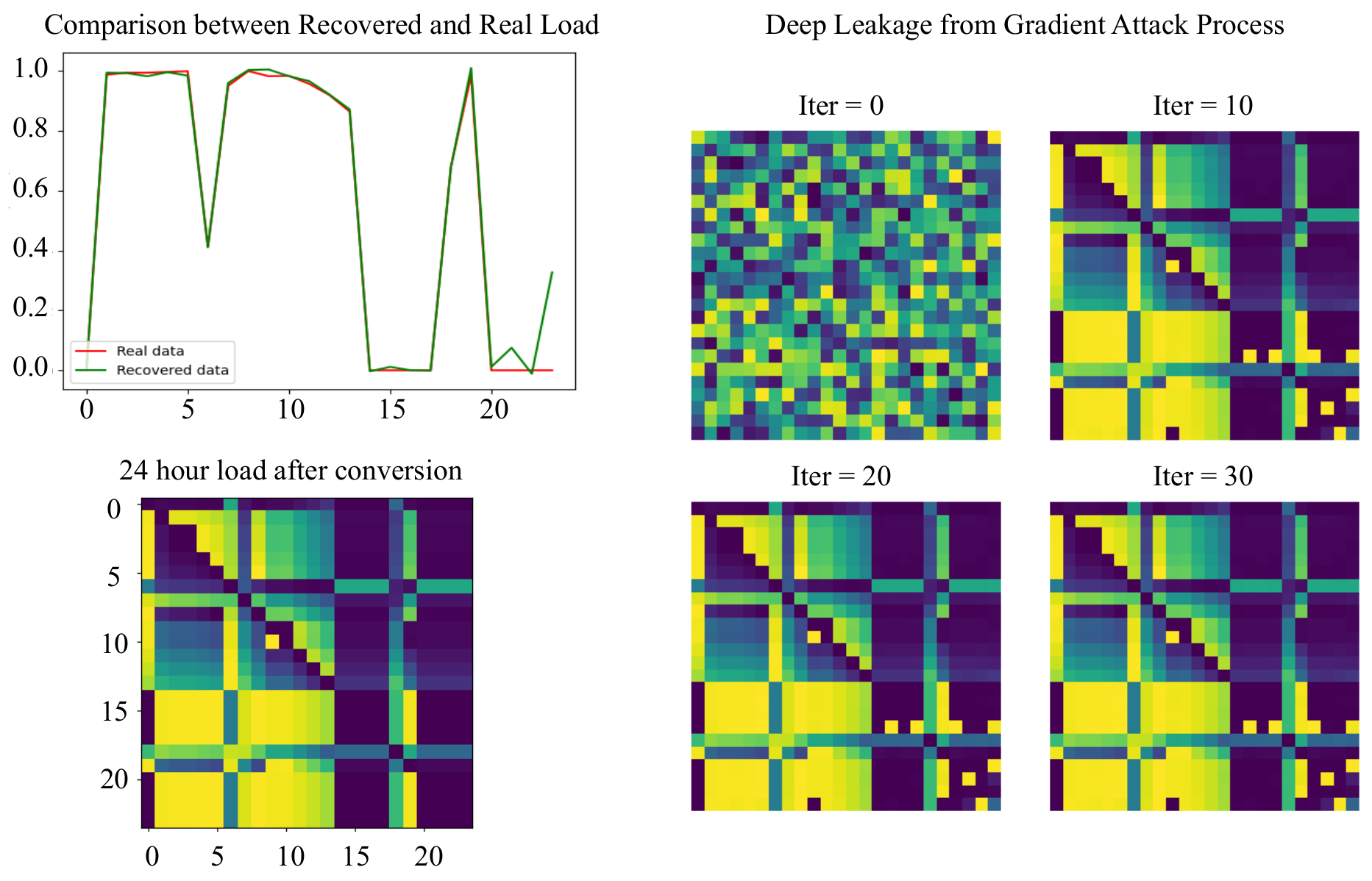}
	\caption{Deep leakage from gradient attack.} 
	\label{fig:tu1}
\end{figure}

The above-mentioned \gls{FL} based algorithms claim the advantage of privacy in terms of not sharing the original data. With the gradual success of various privacy attack technologies in different applications, the load data can also be reverted if the attacker can access the gradients from agents \cite{zhu2019deep}. Here, we use a simple example to show the threat of the attack algorithm to the artificial intelligence-based load forecasting algorithms, shown in Fig. \ref{fig:tu1}.


In Fig.~\ref{fig:tu1}, the two figures on the left are the original consumption data and the original image after conversion. The sub-figures on the right are the regenerated images from the attacker after they obtained the original gradients. As we can see, the attacker can accurately regenerate the original data only after 10 iterations. Therefore, any leakage of the gradient can cause a significant threat to customers' data. To prevent information leakage from agents' gradients, Bonawitz \textit{et al.} \cite{bonawitz2017} proposed using \gls{MPC} to aggregate the gradients privately. Multiparty computation is a cryptographic technique that allows a group of mistrustful parties to perform a computation over secret input data. By using  \gls{MPC}, a group of agents can obtain the aggregated value of all their gradients without ever revealing any individual gradient. The ideas presented in \cite{bonawitz2017} were extensively used and developed in further work related to 
\R{secure aggregation for privacy-preserving federated learning, such as \cite{truex2019, byrd2021, safelearn2021, bell2020, turbo-aggregate2021, safefl2023}. Since these works focus on the federated learning setting, the secure aggregation mechanisms rely on the existence of at least one central server. For the works using additive masking \cite{bonawitz2017, bell2020, turbo-aggregate2021}, a single server is sufficient. In \cite{safelearn2021, safefl2023}, the central aggregator is distributed, meaning that at least two servers are required.}


Multiparty computation achieves privacy by distributing the computation between different parties, which then need to communicate with each other according to the chosen \gls{MPC} protocol to obtain the result of the desired calculation without compromising private input data.
The parties running the \gls{MPC} protocol can either be the agents themselves, or external servers that receive the data from the agents already in encrypted form.
However, there are two possible issues: 

1) \textit{Scalability}: performing computations over private data leads to an increase in communication. Setting the \gls{MPC} parties as external servers helps diminish this overhead, but still results in an expanded network, with each agent needing to connect to every \gls{MPC} party.

2) \textit{Single-point failures}: if the central server in a federated learning system experiences a single-point failure, then the entire system may become unavailable until the server is restored. The \gls{MPC} protocol we use does not solve the issue, as it needs all connected parties. \R{ However, due to the nature of our proposed network topology, no single party (or server) will need to remain online during the full training process.}

\R{Besides the \gls{DLG} attacks, the poisoning attack is another common threat towards \gls{FL} models. A poisoning attack involves one or a few malicious clients injecting extreme outliers or tailored data into the training dataset or model updates to undermine or mislead the global model performance~\cite{tolpegin2020data}. Some researchers developed a Byzantine-robust federated learning model against the poisoning attacks in various ways, such as detecting malicious clients based on their impact on the model performance and constructing new loss functions ~\cite{cretu2008casting,feng2014robust,jagielski2018manipulating,steinhardt2017certified}. However, the common Byzantine-robust models cannot always defend the poisoning attacks, particularly when some of the clients conspire together~\cite{fang2020local}. Among the more advanced defenses, there are two significant approaches. ``Romoa" is a robust model aggregation solution to both solo and collusive attacks. With a hybrid similarity measurement method, it can detect the attacks precisely whether it is targeted or untargeted attacks~\cite{mao2021romoa}. ``FLTrust" reserved a small clean dataset as a reference at the server. By cross-validation, the central server gives a trust score to each collected update. It provides a more precise and less computationally costly way to detect malicious clients~\cite{cao2020fltrust}.  }

To overcome the above-mentioned challenges, we borrow the idea from consensus-based distributed optimization \cite{ding2013consensus, zhao2017distributed, dong2019demand}, which has flexibility in connected network graphs and initialization-free advantages. In this paper, we propose a fully decentralized Markovian Switching short-term residential-user load forecasting algorithm. The proposed distributed algorithm employs Markovian Switching topologies \cite{wang2014seeking}, which can randomly select sets of agents to co-train the model, and therefore reduce the communication complexity in theory and accelerate encryption speed. \R{Furthermore, the \gls{DMS} topology is instinct-resilient against poisoning attacks. In the centralized \gls{FL} training, the central server updates the global model with the average updates across the whole client group each round. Even though the malicious clients take a small part of the whole group, the accumulated strewing effect can significantly undermine the global model over multiple rounds. Unlike the centralized \gls{FL} setting, the \gls{DMS} takes a random set of clients each round, which reduces the chance of malicious clients poisoning attack for every round. The above-mentioned cumulative effect will not happen in the \gls{DMS} setting.} 


The major contributions of this work are summarized as follows:

\R{
\begin{enumerate}
    \item A secure and safe distributed algorithm is proposed for short-term load forecasting and its convergence has been theoretically proved. Different from most existing studies, e.g., \cite{nightingale2022effect, yang2023integrated} and references therein, which are extensively concentrated on federated algorithms, the proposed algorithm in this paper is based on fully \gls{P2P} distributed consensus learning without a central server. 
    \item {The \gls{DMS} is developed for the proposed framework. Different from the traditional FL, the \gls{DMS} has inherited robustness towards both poisoning attacks and \gls{DLG} attacks. Besides that, the \gls{DMS} topology shows advantages in efficiency compared with other distributed learning topologies.}
    \item {We apply the \gls{SecAgg} to the proposed distributed learning topologies. The \gls{SecAgg} ensures no original gradients are shared. Therefore it guarantees data privacy from \gls{DLG} attacks by design. Additionally, we analyzed its impact on the original complexity as an extra add-up layer.} 
    \item {The proposed algorithm has been successfully validated on a real power system load forecasting dataset. Furthermore, the reproducibility and replicability of our work are guaranteed since all source code is available on GitHub for open access\footnote{\url{https://github.com/YingjieWangTony/FL-DL.git}}.}
\end{enumerate}
}

The rest of this paper is organized as follows. The mathematical preliminaries used in this paper are summarized and the researched problem is formulated in Section \ref{sec_pre}. A secure-aggregated and Markovian switching distributed framework for \gls{STLF} is proposed and analyzed in Section \ref{sec_algorithm}. Simulation results and corresponding analysis are presented in Section \ref{sec_sim}. Finally, Section \ref{sec_con} concludes this paper.

\section{Preliminaries}\label{sec_pre}
In this section, we recall some basic concepts (federated learning and distributed learning) and preliminaries related to the proposed \gls{STLF} model, including graph theory and secure aggregation.

\subsection{Federated Learning}
\begin{figure}[htbp]
        \centering
            \includegraphics[width=\linewidth]{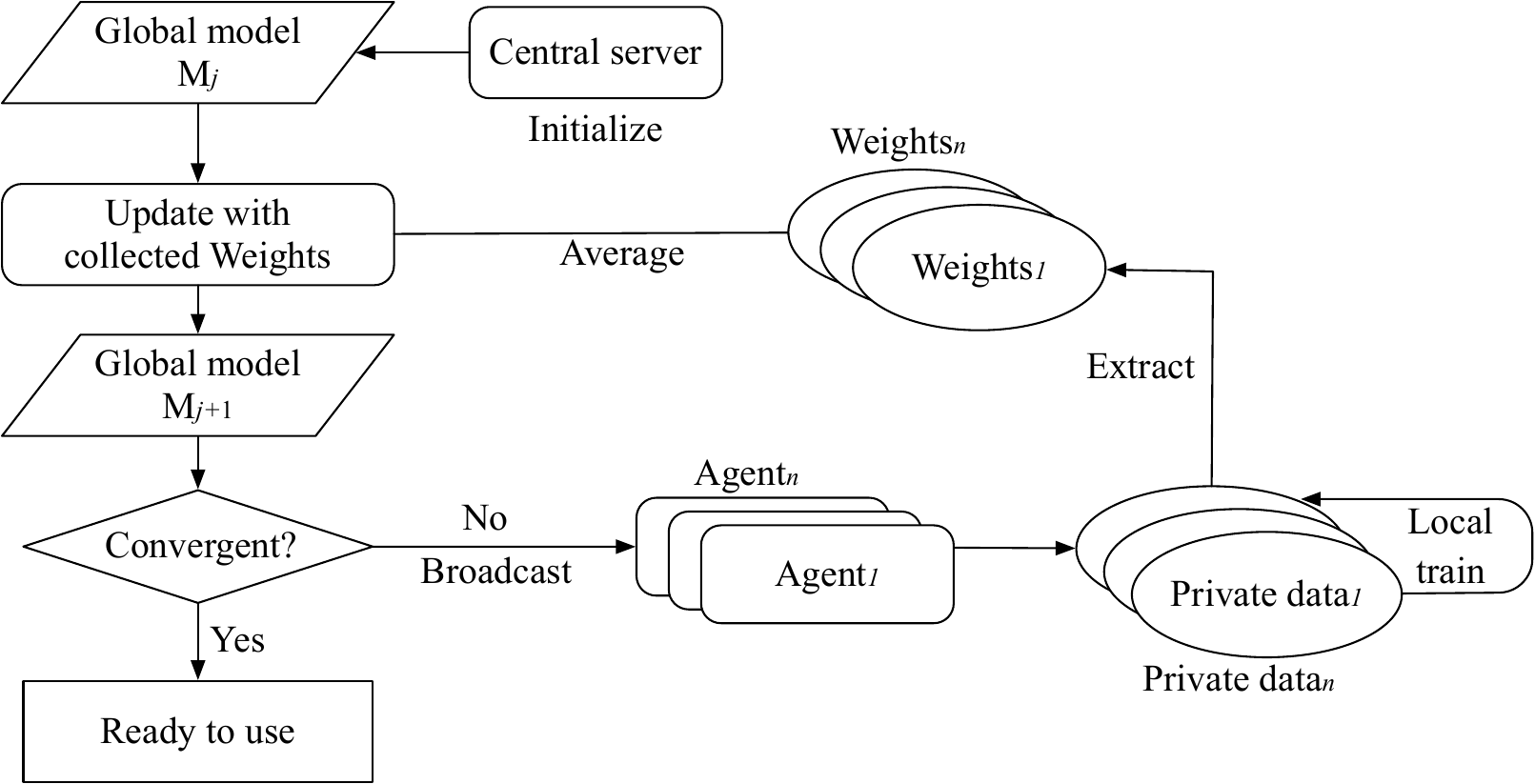}
            \caption{Training procedure of federated learning.}
            \label{fig:2a}
\end{figure}

\gls{FL} is a distributed training framework for neural networks. It decouples the training process and the requirement for centralized data storage. Instead of collecting raw data from the individual agent, the raw data stays locally with agents. The agents use their own data to train the global model, $M_j$. Then, the central server collects the weights updates from the agents. Based on the selected algorithm, for example the \gls{FedAvg}, the server will average the submitted weights and use them to update the global model to $M_{j+1}$. In this way, agents can participate in the model training without revealing their raw data. Figure~\ref{fig:2a} shows the procedure of FL.


In the \gls{FL} framework, the submitted local model update can be ``gradients" or ``weights". It depends on the communication bandwidth, but they are practically the same. In the privacy analysis, we use the term ``weights" for consistency.

\subsection{Graph Theory}
In this paper, it is assumed that the information can be communicated between different computing agents. 
Therefore, the distributed load forecasting conforms to the characteristics of the undirected graph, which can enhance the stability of communicated graphs and reduce the convergence time \cite{ren2008distributed}.

Let $V = \{1,...,N\}$ be a set of $N$ local agents.  Assume that, at any time $t$, the agents constitute a graph ${\cal G}^t = (V^t, E^t)$, where $V^t\subseteq V$ is a set of agents and $E^t\subseteq V^t\times V^t$ is a set of edges. Whenever $(i,j)\in E^t$, we say that they are neighbors. We write $\Mat{L}^t$ as the Laplace matrix of  ${\cal G}^t$. 

\begin{equation}
	\Mat{L}^t_{ij}=\left\{
	\begin{array}{lr}
		-a_{ij}, & i\neq j \\
		\sum_{i\neq j}a_{ij},&i= j 
	\end{array}
	\right.
\end{equation}

To enhance privacy and safety during the training process, the communication graph in this paper is considered as a time-variance topology, called Markovian switching network topology, which means that the graph ${\cal G}^t$ might change over time due to the addition and deletion of agents, which forms a temporal graph ${\cal G}=\{{\cal G}^t\}_{t=0,1,...}$. 
The Markovian switching network has $q$ substructures. 
The transition probability matrix is $\mathbb{T}=[\mathcal{T}_{ij}]\in \mathbb{R}^{q\times q}$, where $\mathcal{T}_{ij}=P\{{\cal G}^{t+1}=j|{\cal G}^t=i\}, \forall i,j\in Q$ with $Q$ being the graph structure topology set.

\subsection{Distributed Learning}
Different from the \gls{FL} framework, the distributed learning algorithm in this paper is designed in a fully decentralized setting, where the agents can only communicate with their adjacent neighbors. 

From \cite{li2021consensus}, the training process of the distributed learning algorithm can be summarized as:
\begin{gather}\label{eqn:localtraining}
\phi_{i}(k) = \theta_i(k) - \gamma \nabla_{\theta_i} {\cal L}(\mathcal{D}_i,M_{PF,i}(\theta_i(k)))\\
\label{eqn:consensusalgorithm}
\theta_i(k+1) = \alpha \sum_{j\in\mathcal{N}}a_{ij}\phi_j(k)
\end{gather}
where $ \theta_{i}(k) $ and $\phi_i(k)  $ denote the weights of the neural network before and after the $ k $th training iteration,  $ \gamma,\alpha >0 $ are the learning rate, and $ \nabla (\cdot)$ is the gradient term regarding empirical risk $\cal L$, which is a measure of how well a machine learning model performs on a given dataset. It is defined as the average of the loss function (i.e., the discrepancy between predicted values and actual values) over the entire dataset. $\mathcal{D}_i$ and $M_{PF,i}$ denote the dataset and local model of $i$th agent, respectively.

Notice that the equations \eqref{eqn:localtraining} and \eqref{eqn:consensusalgorithm} are also called an \gls{LTC} algorithm, where we can easily swap the equations \eqref{eqn:localtraining} and \eqref{eqn:consensusalgorithm}  to get the twinned \gls{CTL} algorithm. 

\subsection{Secure Aggregation}
 
Secure aggregation is achieved through a cryptographic technique, called \gls{MPC}, which allows us to perform calculations over encrypted data. There are different types of \gls{MPC} protocols, with different security guarantees. In this work, we use SCALE-MAMBA \cite{SCALE}, a software framework that implements \gls{MPC} protocols based on secret sharing which have active security, meaning that the protocol remains secure even if up to a certain percentage of the parties performing the computation behaves maliciously and deviates arbitrarily from the protocol. Additionally, the protocols in SCALE-MAMBA are in the preprocessing model. In this model, there is an offline phase, where input-independent data is generated. This can be done at any time, including before the inputs for the desired computation are even available. Later, during the online phase, we can perform our computation faster by using the preprocessed data. 

In this paper, we consider an \gls{MPC} protocol based on the Shamir secret sharing scheme. In this scheme, in order to share a secret value $s$ with a set of $\nu$ parties, the person holding $s$ must generate a secret polynomial $f(X)$ of degree $h$ such that $f(0) = s$. By randomly generating the other polynomial coefficients $f_1, ..., f_h$, the following polynomial is obtained:
\begin{equation*}
    f(X) = s + f_1\cdot X + ... + f_h \cdot X^h.
\end{equation*}
Then, each party $i$ is given a secret share $s_i = f(i)$.
Note that each secret share by itself looks random, but when at least $h+1$ parties combine their shares then the original secret value can be retrieved via Lagrange interpolation.
The choice of degree $h$ will determine how many dishonest parties we allow without compromising the secret since any set of $h$ or fewer parties cannot reconstruct the polynomial. However, to be able to perform all the operations correctly with this scheme, it is also required that a majority of parties is honest.

We write $\share{x}$ when we refer to a secret shared value $x$, and each secret share $i$ is represented as $\share{x}_i$.
All these values are elements of a finite field $\mathbb{F}_p$, where $p$ is a prime number, and operations between secret shared values also take place in $\mathbb{F}_p$.

\begin{figure}[h!]
    \centering
    \begin{subfigure}[t]{\columnwidth}
        \centering
        \includegraphics[width=0.5\linewidth]{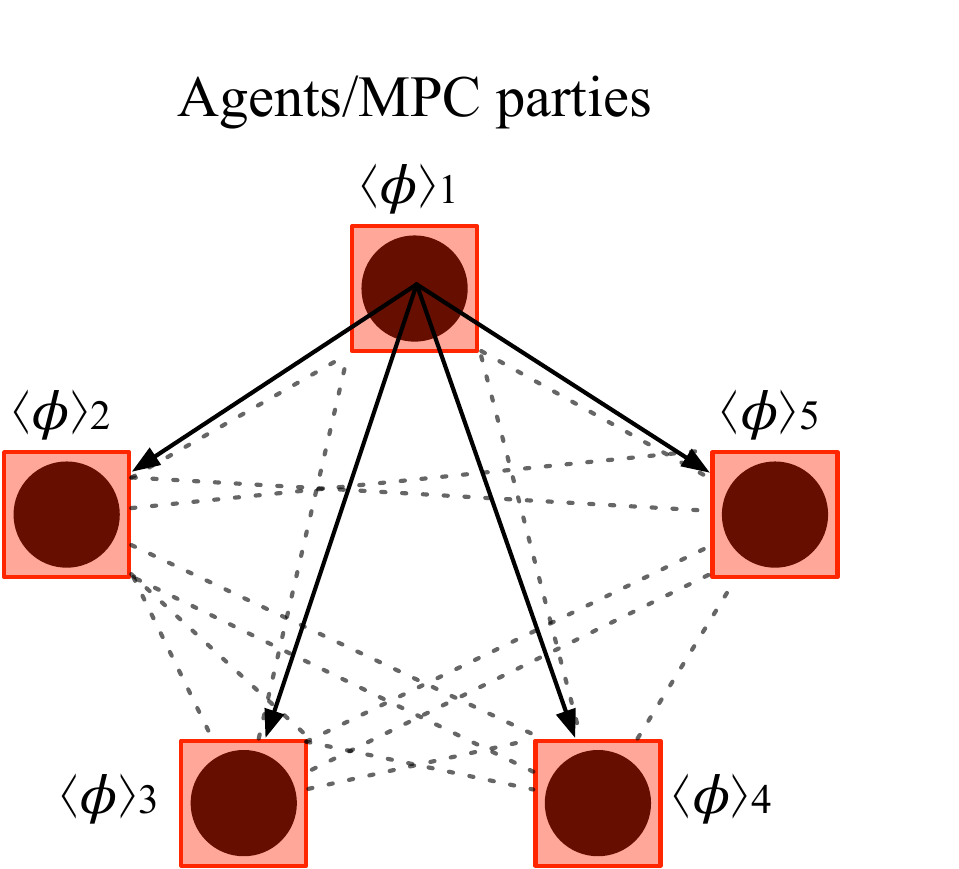}
        \caption{\gls{MPC} parties played by the agents.}
    \end{subfigure}%
    
    \begin{subfigure}[t]{\columnwidth}
        \centering
        \includegraphics[width=0.5\linewidth]{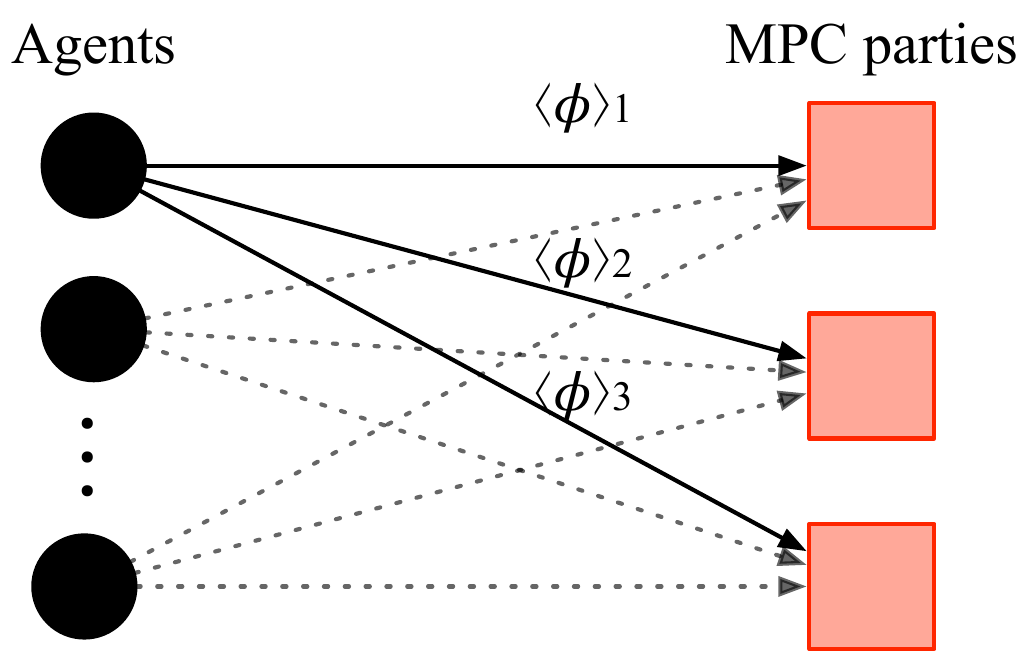}
        \caption{\gls{MPC} parties played by external servers.}
    \end{subfigure}
    \caption{Agent secret sharing a gradient $\phi$ with the \gls{MPC} parties.}
    \label{fig:secretsharing}
\end{figure}

To perform linear operations, i.e., adding two secret shared values or multiplying a secret shared value by a public scalar in $\mathbb{F}_p$, it is enough for each party to locally perform these operations on their own shares. E.g., to calculate $\share{z} = \share{x} + a\cdot\share{y}$, each party $i$ should compute $\share{z}_i = \share{x}_i + a\cdot\share{y}_i$. Multiplying two secret shared values or performing more complex operations such as comparisons requires communication between the parties running the \gls{MPC} protocol so that the correct secret shared result can be reached without revealing the inputs. 
Indeed, communication is generally the bottleneck for computations with \gls{MPC}. However, for secure aggregation only addition is needed, and so communication is only required for the secret sharing of the inputs and reconstructing the final aggregated results. 

As illustrated in Fig. \ref{fig:secretsharing}, it is possible to have agents also act as an \gls{MPC} party, in which case each agent needs to secretly share its own gradients with the other agents (who are also acting as \gls{MPC} parties). In order to improve scalability, we can instead have external servers play the role of \gls{MPC} parties. In this case, we can have as few as three \gls{MPC} parties, to whom the agents will send secret shares of their gradients. These external servers should be run by entities with conflicting interests, in order to avoid collusion and therefore maintain the privacy of the gradients. Note that we should always aggregate more than two gradients in each round, since otherwise the two agents whose weights were aggregated will be able to derive each other's gradients from the revealed aggregated value.

\begin{remark}
\R{We note that our use of Secure Aggregation is not intended to be an improvement upon previous work in the federated learning setting. Indeed, this work is in a fundamentally different setting due to our proposed network topology. Although some existing papers already analyze the use of different topologies for improving the efficiency of secure aggregation \cite{bell2020, turbo-aggregate2021}, they always assume the existence of a central aggregation server.
Additionally, \gls{DP} has also been considered in previous works to ensure that even the final aggregated result does not leak any information \cite{byrd2021, truex2019}. We do not analyze the use \gls{DP} since it is not directly affected by the choice of topology. Nonetheless, it would be relatively straightforward to adapt previous techniques \cite{bohler2021} for combining \gls{MPC} and \gls{DP} to our use case.}
\end{remark}

\section{Distributed Markovian Switching Algorithm}\label{sec_algorithm}
\subsection{Problem Formulation}
In this paper, we focus on the privacy-preserving residential short-term load forecasting problem, where every agent $i$ holds a private dataset $\mathcal{D}_i$, and we define $\mathcal{D}=\mathcal{D}_1\cup ...\cup \mathcal{D}_N$. The $N$ agents will communicate and train a model $M_{DL}$ where $DL$ stands for distributed learning. This will be compared with a federated learning model $M_{FL}$ (i.e., the server collects and modifies the weights from all local agents during the training) and a model $M_C$ trained by centralized learning (i.e., the server collects all raw data from the agents before training).

\subsection{\gls{DMS} Learning}\label{sec:declearn}
The proposed short-term load forecasting algorithm is to utilize decentralized distributed learning to collectively aggregate the local gradients. In a fully decentralized setting,  the agents can only send/collect messages to/from their neighbors but have a collective objective of training a global model $M_{DL}$, which is required to be high-performing and privacy-preserving.

\begin{figure}[htbp]
    \centering
    \includegraphics[width=0.8\columnwidth]{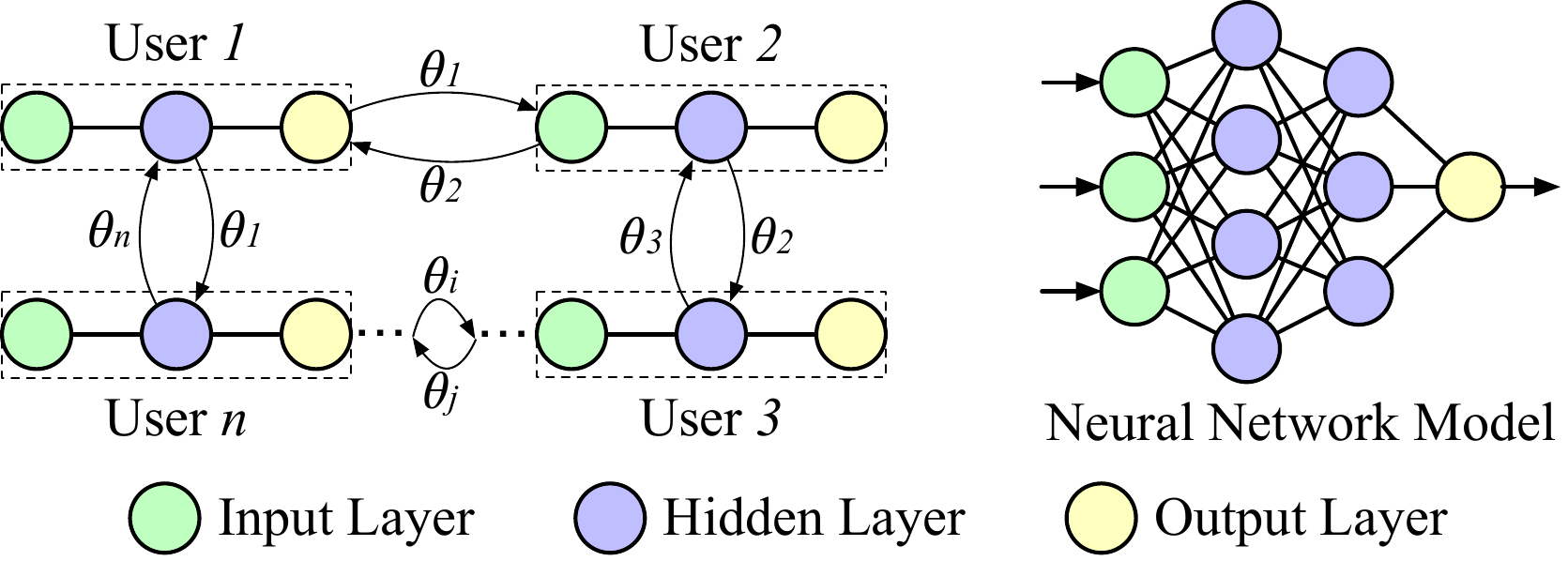}
    \caption{An illustrative framework of distributed learning.}
    \label{DL}
\end{figure}

Every distributed agent is regarded as a self-training agent and cooperates with each other using distributed neural networks with a neurodynamic consensus-based approach, as shown in Fig.~\ref{DL}. The framework consists of a group of neural networks distributed over a connected graph, where the update of coefficients includes two stages: in the first stage, the $i$th agent trains its neural network locally using the dataset $\mathcal{D}_{i}$, and in the second stage, the consensus of their locally trained weights is performed.  With this setting, all the agents are able to obtain the same neural network if every sub-dataset is available to at least one local agent, which will be demonstrated by the convergence analysis.

Minimizing the empirical risk allows the model to generalize well to new, unseen data. The idea is that if the model is trained on a dataset that is representative of the real-world data it will encounter in the future, then minimizing the empirical risk should result in a model that can make accurate predictions on new data. Therefore, the objective of the network is to minimize the empirical risk over the entire dataset, given by 
\begin{align}\label{empiricalrisk}
\min {\cal L}(\mathcal{D},M_{PF}(\theta)) = \sum_{i=1}^{N} {\cal L}(\mathcal{D}_i,M_{PF,i}(\theta_i))
\end{align}
where $ \mathcal{D} $ and $ \mathcal{D}_i $ denote the entire dataset and the sub-dataset of agent $ i $, respectively; $ \theta $ and $ \theta_i $ are the weights of the global and local neural networks; $ M_{PF} $ and $ M_{PF,i}$ represent the models of the global neural network and the local ones, respectively.
It is worth noting that the local learning process, i.e., 
\begin{align}\label{eqn: local empirical risk}
\min  {\cal L}(\mathcal{D}_i,M_{PF,i}(\theta_i))
\end{align} 
does not necessarily yield the minimization of the global empirical risk in (\ref{empiricalrisk}), since the models $ M_{PF,i} $ are different. Consequently, this necessitates the employment of consensus tools in cooperation with the training process.

Note that the sub-dataset of each agent is assumed to be stationary. For the $ i $th agent with $ n_i $ samples in $ \mathcal {D}_i $, the gradient can be calculated by 
\begin{align}
\nabla_{\theta_i} {\cal L}(\mathcal{D}_i, M_{PF,i}(\theta_i)) =\frac1 {n_i} \sum_{k=1}^{n_i} \nabla_{\theta_i} \ell(s^{k},\theta_i)
\end{align}
where $ s^k $ represents the $ k $th data sample in $ \mathcal{D}_i $, and $ \ell(s^k,\theta_i) $ is the empirical risk of sample $ s^k $ on a model with weights $\theta_i$. 
Similarly, the global gradient is given by 
\begin{align}
\nabla_\theta {\cal L}(\mathcal D, M_{PF}(\theta)) = \frac 1 n \sum_{k=1}^{n} \nabla_{\theta} \ell(s^{k},\theta)
\end{align}
where $ n = \sum_{i=1}^N n_i $ is the total number of samples in $ \mathcal{D} $.

Gradient-based methods have been widely used in training neural networks, due to their high efficiency and flexibility \cite{bottou2007tradeoffs,tang2018d,saad1998online}. In this paper, a gradient descent algorithm for agent $i$ is formulated as 
\begin{align}\label{localtraining}
\phi_{i}(k) = \theta_i(k) - \gamma \nabla_{\theta_i} {\cal L}(\mathcal{D}_i,M_{PF,i}(\theta_i(k))) 
\end{align}
\begin{align}\label{consensusalgorithm}
\share{\theta_i(k+1)} = \alpha \sum_{j\in\mathcal{N}}a_{ij}\share{\phi_j(k)}
\end{align}
where $ 0\leq\alpha\leq1 $ is positive constant;
$a_{ij}$ denotes the communication link between agent $i$ and agent $j$; $a_{ij}>0$ if they can communicate, otherwise, $a_{ij}=0$. Note that for each agent $ i $, the consensus only requires the updated weights $ \phi_{j} $ from its adjacent neighbors $ j\in \mathcal{N}_i $ with $\mathcal{N}_i$ denoting the neighbors of agent $i$. Recall also that we add the updated weights using secure aggregation, hence the secret share notation used in Equation \ref{consensusalgorithm}. The shares of $\share{\theta_i(k+1)}$ will then be sent by the \gls{MPC} parties to agent $i$, who can finally reconstruct the value $\theta_i(k+1)$.

Based on the above discussion, the overall structure of the fully decentralized distributed algorithm for a region can be summarized in Algorithm~\ref{DLS}.

\begin{algorithm}
	\textbf{Initialization:}\\
	\ \ \  for each agent $ i \in \mathcal V $: \\
	1. initialize the weights of local neural network $ \theta_i $;\\
	2. set the required consensus speed by changing the coefficients in Equation (\ref{consensusalgorithm}).
	
	\textbf{Iteration:}\\
	3. set $ k:=k+1 $, for $ i\in \mathcal V $ \\
	4. train the local neural network using the gradient descent method (\ref{localtraining}) with locally stored dataset $ \mathcal D_i $ to obtain $ \phi_i(k) $;\\
        5. apply secure aggregation to encrypt original weights $\share{\phi_i(k)}$; \\
	6. run the consensus operation (\ref{consensusalgorithm}) to drive the globally averaged weights $ \theta_i(k+1) $ by  communicating with its neighbors $ \mathcal N_i $;  \\
	\textbf{Termination:} termination condition is satisfied or iteration budget is reached. 
	\caption{Distributed Markovian Switching (DMS)} \label{DLS}
\end{algorithm}

\subsection{Correctness Analysis on Performance/Convergence 
} \label{sec_alg_performance}

We need to show that Algorithm~\ref{DLS} is able to converge with all agents having the same model. 
Considering the generality, we choose \gls{MSE} as the loss function.

\RR{MSE is a widely used measure in statistics and machine learning for quantifying the difference between estimated values and the actual values \cite{bickel2015mathematical}. It is calculated as the average of the squares of the errors or deviations, that is, the difference between the estimator and what is estimated. Mathematically, MSE is defined as
    \begin{equation}
    MSE = \frac{1}{n_s} \sum_{i=1}^{n_s}(y_i - \hat{y})^2
\end{equation}
where $n_s$ is the number of samples, $y_i$ and $\hat{y_i}$ are the true and predicted values, respectively.}

First of all, two assumptions are needed.
With these two assumptions and the mean value theorem, we can obtain the closed-loop \gls{MSE} dynamics, and therefore the convergence of the \gls{MSE} can be proved with a designed learning rate.
\begin{assumption}\label{assumption1}
The Hessian matrix of local empirical risk is bounded, i.e.,:
\begin{equation}\label{eq:assumption1}
    \underline{p}_i\Mat{I}_n \leq \nabla^2{\cal L}(\mathcal{D}_i,M_{PF,i}(\theta_i))\leq \overline{p}_i\Mat{I}_n
\end{equation}
where $0\leq\underline{p}_i\leq\overline{p}_i$ are non-negative constants.
\end{assumption}

\begin{assumption}\label{assumption2}
The gradient noise, $w_i$, satisfies the following properties i.e.,: 
\begin{align}
    \mathbb{E}[w_i] &= 0 \\
    \mathbb{E}[\|w_i\|^2] &\leq \xi_i^2
\end{align}
where $\xi_i$ is non-negative constant, $\|\cdot\|$ denotes the L2-norm of the argument. 
\end{assumption}
\begin{remark}
Assumption 1 indicates that the network objective, $\sum_{i=1}^{N} {\cal L}(\mathcal{D}_i,M_{PF,i}(\theta_i))$, is strongly convex, which means that the neural network has a unique and optimal solution/weights $\theta^*$. In distributed optimization problems, some assumptions have been made, including the use of bounded Hessian matrices for convergence analysis. Other works \cite{nedic2009distributed,deng2017distributed,zhu2018continuous} have used bounded gradients with set constraints, which is not feasible for unconstrained problems. Our assumption is less restrictive because we can solve problems with unbounded gradients. \\
Assumption 2 implies that the noise, $w_i$, is unbiased and has uniformly bounded variance $\xi_i^2$, which is a stronger assumption and has been considered in many studies \cite{duchi2011dual,tang2018d}.
\end{remark}

\begin{theorem}\label{theorem:1}
Under Assumptions 1 and 2, if the learning rates satisfy $0\leq\gamma_i\leq\frac{2}{\overline{p}_i}$,
the worst \gls{MSE}, $\|\mathbb{N}\|^\infty$ in the distributed learning network converges to
\begin{equation}\label{eq13}
    \lim_{k\rightarrow\infty}\|\mathbb{N}(k)\|^\infty\preceq\alpha\Xi
\end{equation}
\end{theorem}
where $\gamma_i$ is the local learning rate for $i$th agent. $\overline{p}_i$ is the local upper bound of the Hessian matrix as shown in equation \eqref{eq:assumption1}. $\mathbb{N}$ is the \gls{MSE} vectors of the distributed neural networks, which is defined later in equation \eqref{eqb:21}. $\Xi = diag(\xi_1^2,\cdots,\xi_N^2)$ is the matrix format of the uniformly bounded variances. 

\RR{
\begin{remark}
    In Theorem \ref{theorem:1}, $0\leq\gamma_i\leq\frac{2}{\overline{p}_i}$ provides an upper bound for designing the learning rates, based on which we can prove the convergence of the proposed algorithms. Equation \eqref{eq13} reveals how the learning rates, the bounds of the Hessian matrix, and the noise variance will affect the performance of the algorithms. It means that minimizing the learning rates can lead to a reduction in the optimality error. The designer can balance between the convergence speed and the optimality error according to the real applications by adjusting the heterogeneous learning rates. 
    In other words, this theorem serves as a reference to select learning rates during our training process in both DL and FL frameworks.
    In \eqref{eq13}, we have 3 parameters, $\overline{p}_i, \alpha$ and $\xi_i$. $\overline{p}_i$ can be obtained from the loss function, $\alpha \leq 1$ is a fixed parameter, and the coefficient of gradient noises $\xi_i$ should be approximated according to the practical applications.
\end{remark}
}

\begin{proof}

To simplify the notations in the analysis process below, we rewrite the weights in a compact form by integrating all weights into a vector form as: 
\begin{align}
    \boldsymbol{\Phi}(k) &= \text{col}(\phi_1(k),\cdots,\phi_N(k))\\
    \boldsymbol{\Theta}(k) &= \text{col}(\theta_1(k),\cdots,\theta_N(k))
\end{align}
Then, we can rewrite the proposed distributed algorithm \eqref{localtraining} and \eqref{consensusalgorithm} in the following form:
\begin{align}\label{eq:14}
    \boldsymbol{\Phi}(k) &= \boldsymbol{\Theta}(k) - (\Gamma\otimes \Mat{I}_n)\boldsymbol{\nabla}{\cal L}(\boldsymbol{\Theta}(k)) + \boldsymbol{\Omega}(k)\\
    \boldsymbol{\Theta}(k+1) &= \alpha(\mathcal{A}\otimes \Mat{I}_n)\boldsymbol{\Phi}(k)\label{eq:15}
\end{align}
where $\Gamma = diag(\eta_1,\cdots,\eta_N)$, $\boldsymbol{\nabla}{\cal L}(\boldsymbol{\Theta}) = col(\nabla {\cal L}_1(\boldsymbol{\Theta}),\cdots,\nabla {\cal L}_N(\boldsymbol{\Theta}))$, $\boldsymbol{\Omega} = col(\omega_1,\cdots,\omega_N)$. $\mathcal{A}$ is the Laplacian matrix, which is defined as $\mathcal{A} = [a_{ij}]_{N\times N}$. $a_{ij}>0$ if the agent $i$ is communicating with agent $j$. Note that each row of the $\mathcal{A}$ sums to $1$, which is also called row-stochastic matrix \cite{ren2008distributed}.

To confirm that the \gls{MSE} converges, we transfer the weights to an error recursion formation:
\begin{align}
    \hat{\boldsymbol{\Phi}}(k) &= \boldsymbol{\Phi}(k) - \theta^*\textbf{1}_N\\
    \hat{\boldsymbol{\Theta}}(k) &= \boldsymbol{\Theta}(k) - \theta^*\textbf{1}_N
\end{align}

Furthermore, to relate the gradient term $\boldsymbol{\nabla}{\cal L}(\boldsymbol{\Theta}(k))$ with error term $\hat{\boldsymbol{\Theta}}(k)$, we have the mean value theorem from \cite{rudin1976principles}:
\begin{equation}\label{mvt}
    \nabla_z g(y) = \nabla_z g(x) + [\int_0^1\nabla_z^2g[x+\tau(y-x)]d\tau](y-x)
\end{equation}

Substituting the $\boldsymbol{\Theta}(k)$ and $\theta^*\textbf{1}_N$ into equation \eqref{mvt} as $y$ and $x$:
\begin{equation}\footnotesize
    \boldsymbol{\nabla}_\theta {\cal L}(\boldsymbol{\Theta}(k)) = \boldsymbol{\nabla}_\theta {\cal L}(\theta^*\textbf{1}_N) + [\int_0^1\nabla_\theta\boldsymbol{\nabla} {\cal L}[\theta^*\textbf{1}_N+\tau(\hat{\boldsymbol{\Theta}}(k))]d\tau](\hat{\boldsymbol{\Theta}}(k))
\end{equation}

Since $\boldsymbol{\nabla}_\theta {\cal L}(\theta^*\textbf{1}_N) = 0$, and we let $P(k) = \int_0^1\nabla_\theta\boldsymbol{\nabla}{\cal L}[\theta^*\textbf{1}_N+\tau(\hat{\boldsymbol{\Theta}}(k))]d\tau$. The equation \eqref{eq:14} can be rewritten as:
\begin{equation}\label{eq:18}
    \hat{\boldsymbol{\Phi}}(k) = \hat{\boldsymbol{\Theta}}(k) - (\Gamma\otimes \Mat{I}_N)P(k)\hat{\boldsymbol{\Theta}}(k)+\boldsymbol{\Omega}(k)
\end{equation}

Combining \eqref{eq:15} and \eqref{eq:18} together, we can get the error term update law as:
\begin{equation}\small
    \hat{\boldsymbol{\Theta}}(k+1) = \alpha(\mathcal{A}\otimes \Mat{I}_n)[\Mat{I}_nN - (\Gamma\otimes \Mat{I}_N)P(k)]\hat{\boldsymbol{\Theta}}(k) + \alpha(\mathcal{A}\otimes \Mat{I}_n)\boldsymbol{\Omega}(k)
\end{equation}

We define the \gls{MSE} vectors of the distributed neural networks as:
\begin{align}\label{eqb:21}
    \mathbb{M}(k) =& \text{col}(\mathbb{E}\|\hat{\boldsymbol{\phi}}_1(k)\|^2,\cdots,\mathbb{E}\|\hat{\boldsymbol{\phi}}_N(k)\|^2)\\
    \mathbb{N}(k) =& \text{col}(\mathbb{E}\|\hat{\boldsymbol{\theta}}_1(k)\|^2,\cdots,\mathbb{E}\|\hat{\boldsymbol{\theta}}_N(k)\|^2)
\end{align}

By taking the Euclidean norm for each agent's error dynamics, we have:
\begin{equation}\label{eqb:22}
    \|\hat{\boldsymbol{\theta}}_i(k+1)\|^2 = \|\alpha \sum_{j\in\mathcal{N}}a_{ij}\hat{\boldsymbol{\phi}}_j(k)\|^2
\end{equation}

Since the norm function $\|\cdot\|$ is a convex function, applying Jensen's inequality to \eqref{eqb:22}:
\begin{equation}\label{eqb:23}
    \|\hat{\boldsymbol{\theta}}_i(k+1)\|^2 \leq \alpha \sum_{j\in\mathcal{N}}a_{ij}\|\hat{\boldsymbol{\phi}}_j(k)\|^2
\end{equation}

Taking the expectation of both sides of \eqref{eqb:23}, we can obtain:
\begin{equation}
    \mathbb{N}(k+1)\preceq\alpha\mathcal{A}\mathbb{M}(k)
\end{equation}

Taking the Euclidean norm and expectation of $\hat{\boldsymbol{\Phi}}_i(k)$ leads to:
\begin{equation}
    \mathbb{M}(k) = \mathbb{E}\|[\Mat{I}_{nN} - (\Gamma\otimes \Mat{I}_N)P(k)]\hat{\boldsymbol{\Theta}}(k)\|^2 + \mathbb{E}\|\boldsymbol{\Omega}(k)\|^2
\end{equation}

Based on Assumption 1, the bound of the symmetrical matrix $P(k)$ could be obtained:
\begin{equation}
    \underline{p}_i\Mat{I}_n \leq P(k)\leq \overline{p}_i\Mat{I}_n
\end{equation}
and therefore:
\begin{equation}
    0 \leq [\Mat{I}_{nN} - (\Gamma\otimes \Mat{I}_N)P(k)]\leq \overline{\lambda}_i^2\Mat{I}_n
\end{equation}
where $\overline{\lambda}_i = \max{(1-\gamma_i\underline{p}_i)^2,(1-\gamma_i\overline{p}_i)^2}$.

Based on Assumption 2, we have:

\begin{equation}
    \mathbb{E}\|\boldsymbol{\Omega}(k)\|^2\leq\Xi^2
\end{equation}

Until now, the closed-loop \gls{MSE} error can be derived as:
\begin{equation}
    \mathbb{N}(k+1)\preceq\alpha\mathcal{A}\Lambda\mathbb{N}(k)+\alpha\mathcal{A}\Xi
\end{equation}
where $\Lambda = diag(\overline{\lambda}_1,\cdots,\overline{\lambda}_i)$

Considering the worst case \gls{MSE}, $\|\mathbb{N}(k+1)\|^\infty$,

\begin{equation}\label{eqn:final}
    \|\mathbb{N}(k+1)\|^\infty\preceq\alpha\|\mathcal{A}\Lambda\|^\infty\|\mathbb{N}(k)\|^\infty+\alpha\|\mathcal{A}\Xi\|^\infty
\end{equation}

To guarantee the convergence of \eqref{eqn:final}, we must have $\|\mathcal{A}\Lambda\|^\infty \leq 1$. We also have $\|\mathcal{A}\|^\infty = 1$ and $\alpha\leq 1$, so we have
\begin{equation}
    \overline{\lambda}_i\leq 1
\end{equation}
which means:
\begin{align}
    (1-\gamma_i\underline{p}_i)^2&\leq 1\\
    (1-\gamma_i\overline{p}_i)^2&\leq 1
\end{align}
Because of $\underline{p}_i \leq \overline{p}_i$, we should choose the learning rate between the range to guarantee the convergence of the proposed distributed learning algorithm:
\begin{equation}
    0\leq\gamma_i\leq\frac{2}{\overline{p}_i}
\end{equation}

    
\end{proof}

\R{After the convergence analysis, we also provide a theoretical analysis of the proposed algorithm due to the complexity of the algorithm is not easy to visualize. 
We evaluate the complexity based on the communicated components' size with a normalized factor, $O(n|E|)$.
The algorithm is a gradient-based algorithm, which is a widespread technique in machine learning and optimization. Gradient-based algorithms are efficient in terms of computation, as they only require the calculation of the gradient of the objective function, which can be done using simple mathematical operations. This makes the algorithm easy to implement and computationally efficient, which is a desirable feature in many real-world applications.}

\R{In terms of communication complexity, the proposed algorithm only requires the agents to communicate with their immediate neighbors. In other words, each agent only needs to exchange information with the agents that are in its direct vicinity. This limited communication requirement is beneficial as it reduces the overall complexity of the algorithm and makes it more suitable for distributed implementation. Moreover, the proposed algorithm removes the requirement of the central server, which can be a bottleneck in some systems. }

\R{In this setting, the communication complexity for each agent is $O(n|E|)$, where $E$ is the set of communication edges and $n$ is the number of iterations, because in every iteration the agent only needs to send one message to their neighbors and every message is of constant size $O(1)$. The computational complexity for every agent is also $O(n|E|)$, since every iteration the agent can process the received messages and update the local state in a linear way. 
It is noticed that due to the randomly connected characteristic of Markovian switching topology, the communication edges in each round are not a constant value, e.g. $E\in\left[1,30\right], n = 30$. }

\section{Case Studies}\label{sec_sim}
\begin{figure}[htbp]
    \centering
    \includegraphics[width=\columnwidth]{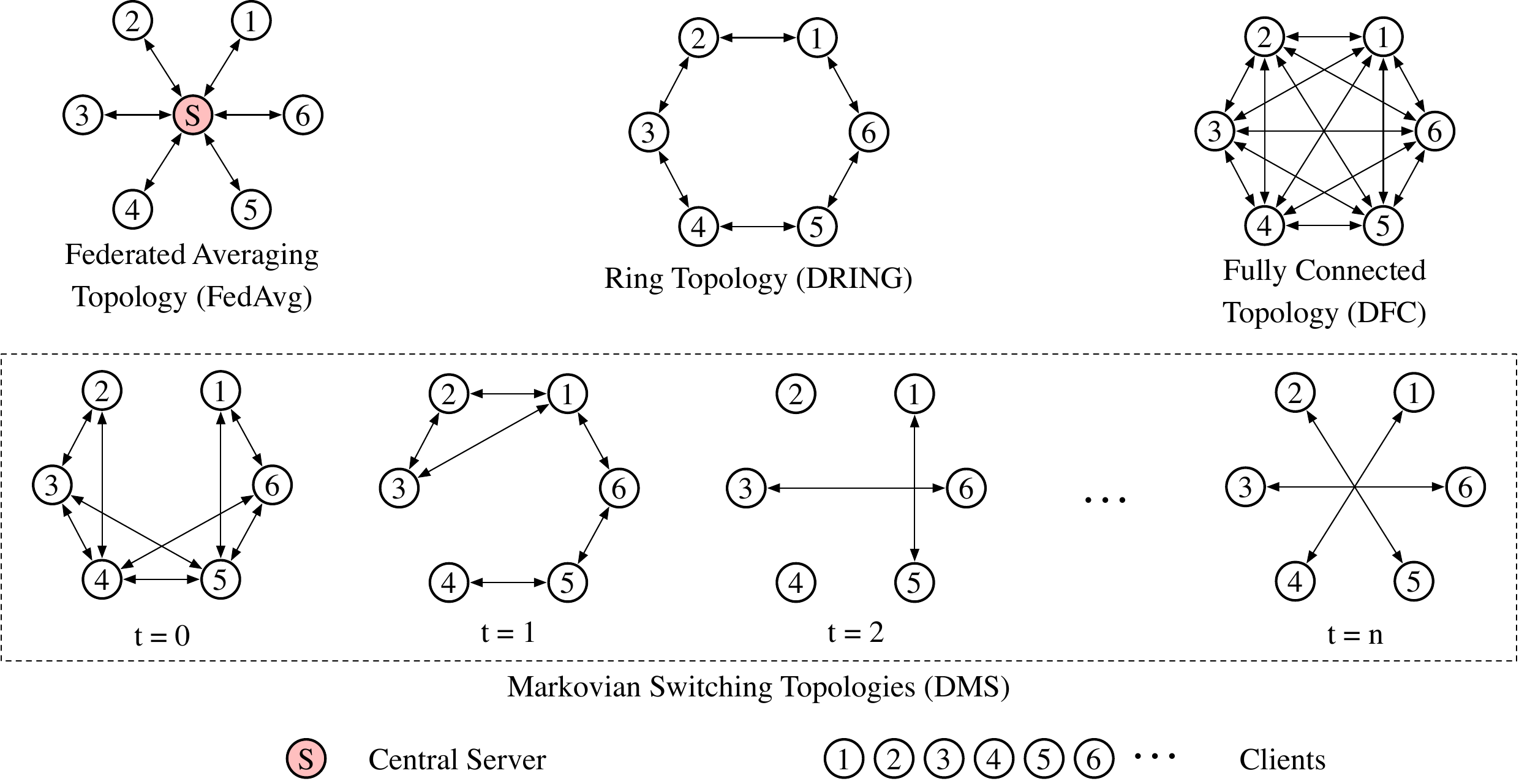}
    \caption{Communication topologies of different strategies.}
    \label{communicationgraphs}
\end{figure}

In our study, we propose a new training strategy called \gls{DMS} learning, which utilizes a dynamic switching mechanism between different topologies during the learning process. To evaluate the performance of \gls{DMS}, we compare it with four other popular training strategies: Centralized Learning, \gls{FedAvg}, \gls{DFC}, and Distributed Learning with Ring Topology (DRING). Here, we use Fig.~\ref{communicationgraphs} to illustrate the communication differences between different learning strategies. These five training strategies are implemented on four state-of-the-art deep learning neural network models: DNN \cite{hippert2001neural}, CNN \cite{liu2018time}, Wavenet \cite{oord2016wavenet}, and LSTM\cite{kong2017short}. The experiments are conducted on a dataset with a large number of samples, and the results are evaluated based on the accuracy, convergence rate, and communication cost of the trained models.
\subsection{Data Sources}
The Smart Metering Electricity Customer Behavior Trials are used as our case study \cite{CER}. It contains over 5000 Irish home and business participants during 2009 and 2010. Their electricity consumption is recorded by smart meters with 30 minutes intervals. The longest record is from January 1st, 2009 to June 30th, 2010.
After data cleaning and clustering, we selected 30 households to present a virtual energy community. The selection was made among 30 houses that were clustered together with all methods and obtained a high score in each.

\begin{figure}[htbp]
    \centering
    \includegraphics[width=\columnwidth]{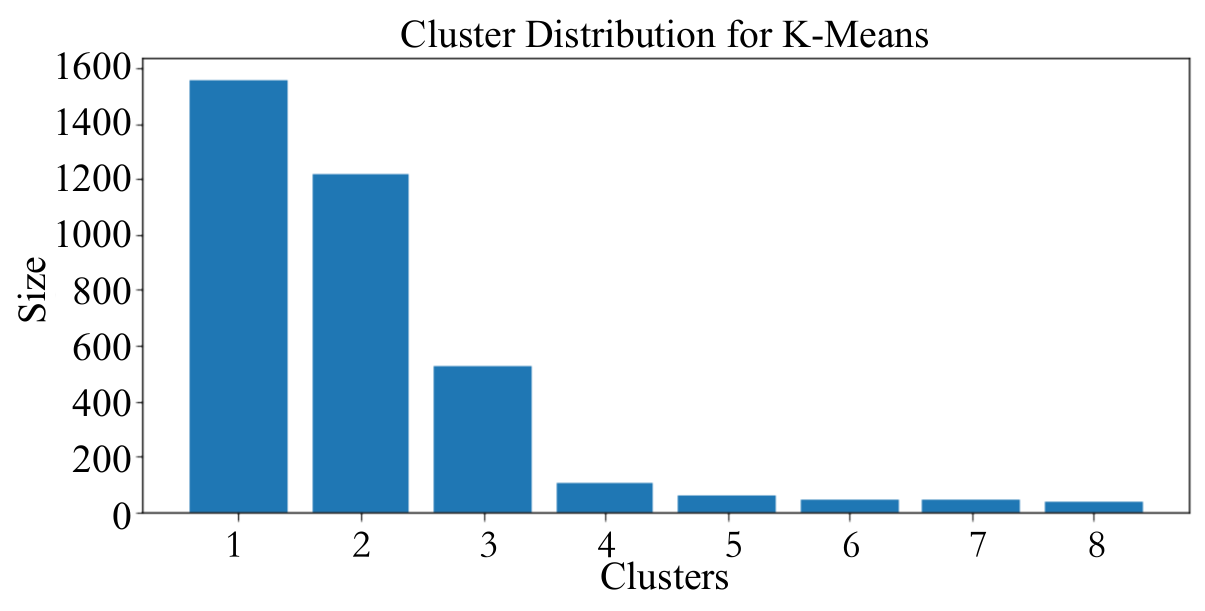}
    \caption{The clustering result with K-means.}
    \label{fig:KMED}
\end{figure}

More specifically, in this case, we use a typical non-Independently Identically Distributed (non-IID) dataset with a large group of agents. It is not realistic and practical to directly apply \gls{FL} to the raw dataset. Therefore, we performed the K-means algorithm to cluster the dataset into small groups \cite{mansour2020three}.
The clustering result can be seen as Fig. \ref{fig:KMED}.

As we can see from Fig. \ref{fig:KMED}, $Cluster_1$ contains most of the agents. It indicates this cluster represents the most common consumption behavior. Therefore, we randomly select 30 agents from $Cluster_1$ to perform our further study in this paper. 

\subsection{Experiment Setup}
The evaluation is based on simulations. 
We programmed the forecasting code in Python and based on the machine learning framework provided by FLOWER\footnote{https://flower.dev/}(flwr), where the neural network models are written in PyTorch\footnote{https://pytorch.org/}.

For the secure aggregation experiments, we use four machines with an Intel i-9900 CPU and 128GB of RAM, and four machines with an Intel Core i7-770 CPU and 32GB of RAM. For experiments with up to 4 \gls{MPC} parties, only the first group of machines is used (one machine per \gls{MPC} party). For experiments with more than 8 parties, each machine runs more than one party.
The ping time between the machines is 1.003 ms.

\subsection{Experiment Results and Analysis}
In this section, we evaluate the performance of the proposed algorithm by assessing it against four key aspects: scalability, privacy, accuracy, and complexity. By evaluating the algorithm from these four key perspectives, we can ensure that it meets the standards required for practical use.

\begin{figure}[htbp]
	\centering
        \includegraphics[width=\columnwidth]{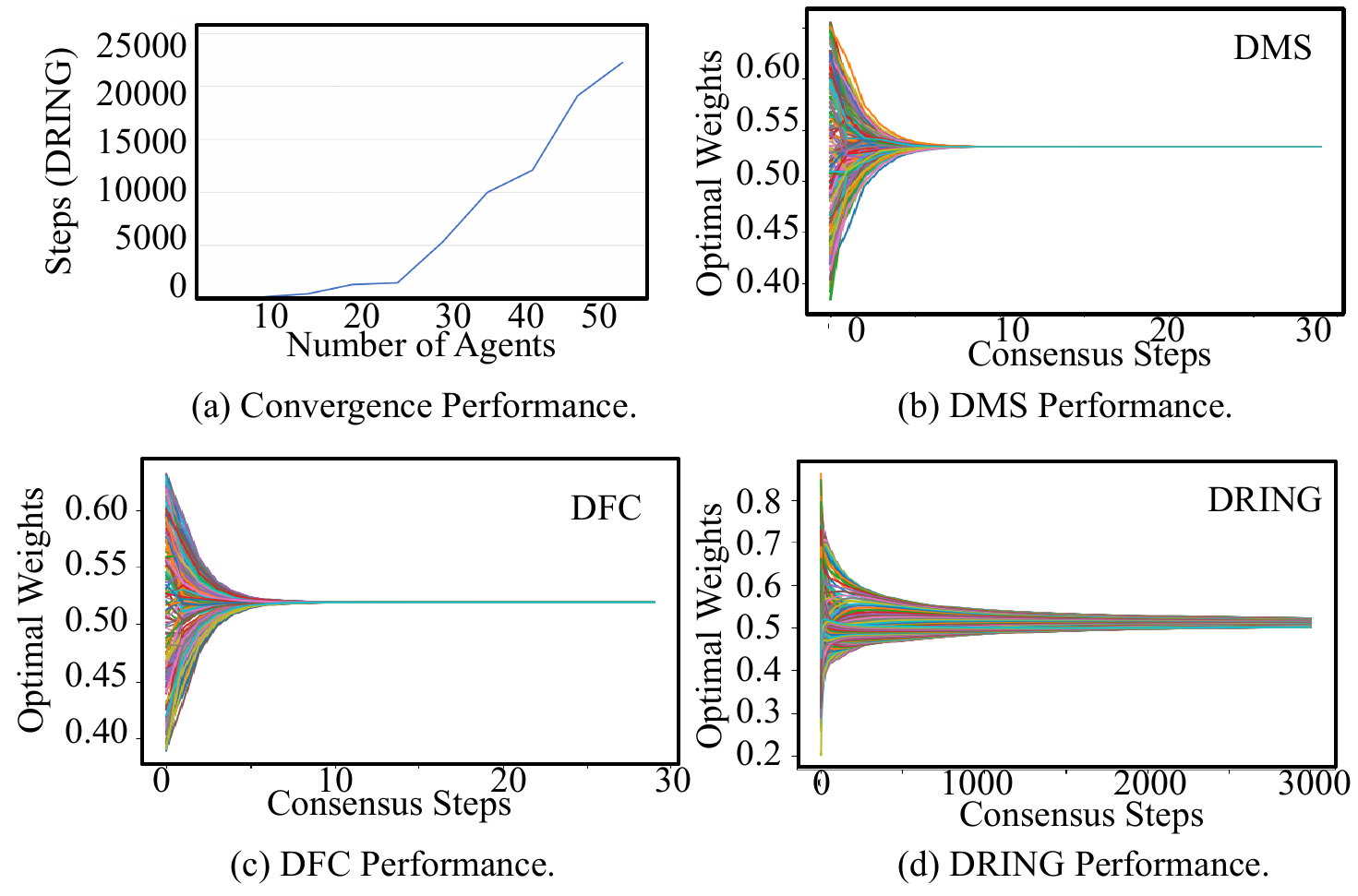}

	\caption{Scalability of different decentralized learning topologies.}\label{fig:scalability}
\end{figure}

\paragraph{Scalability}
We measure the algorithm's ability to handle increasing amounts of data and agents, ensuring that it can scale to meet the needs of large-scale applications. The evaluation method for scalability is the convergence speed.
The proposed algorithm can handle a large number of agents with a relatively small number of neighbors. This allows for more efficient use of resources and makes it possible to use the algorithm in a distributed setting, where the number of agents can be large. 
To demonstrate the scalability of the proposed algorithm, we first analyze the converge steps in the slowest \gls{DRING}. Then, we conducted an experiment to work with up to 200 distributed learning agents and 3 different communication topologies (\gls{DMS}, \gls{DFC} \cite{li2021consensus} and \gls{DRING} \cite{zhao2018consensus}). The experimental results are shown in Fig. \ref{fig:scalability}. 

The experiment was designed to evaluate the effect of increasing the number of agents on the algorithm's convergence performance. Figure~\ref{fig:scalability}(a) shows that the convergence performance increases almost linearly with the number of agents under the \gls{DRING} topology, while the other three figures show the convergence of a 200-agent system with different topologies. This indicates that the proposed algorithm is scalable and can easily adapt to large-scale federated learning systems. Actually, both the communication and computational complexity of an agent do not depend on the set of agents, but rather the set of neighbors. This means that as long as the number of neighbors is kept unchanged, the algorithm can scale to as many agents as possible. This is a significant advantage over other federated learning algorithms that have a high computational and communication complexity with respect to the number of agents. From the simulation results, it can be observed that due to the random communication approach, the \gls{DMS} topology significantly improves convergence speed compared to the \gls{DRING} topology, and achieves convergence speed almost equivalent to that of the \gls{DFC} topology. Theoretically, the \gls{DMS} topology reduces 50\% of the communication edges compared to the \gls{DFC} topology.

\paragraph{Privacy}
For secure aggregation, we used an \gls{MPC} protocol based on Shamir secret sharing which remains secure as long as the majority of the parties are honest (i.e., when there are three parties, at most one can be malicious). If this is satisfied and a dishonest party deviates from the protocol, the honest parties will detect it and abort the protocol with probability $1-1/p$, where $p$ is the order of the field. We choose $p$ such that $\log_2 p = 128$, so that dishonest behavior is caught with overwhelming probability.

We implemented secure aggregation using SCALE-MAMBA and performed experiments for different communication topologies. 
For the \gls{FedAvg} topology, we consider 3 external \gls{MPC} parties, receiving secret shared gradients from the 30 agents in every round.
For the \gls{DRING} topology, each agent acts as an \gls{MPC} party and performs the secure aggregation protocol together with the two connected neighbors.
For the \gls{DFC} topology, there are 30 \gls{MPC} parties played by the agents themselves, who secretly share their gradients with all other 29 agents in every round.
In the \gls{DMS} topology, the \gls{MPC} parties are played by the subset of agents chosen to aggregate their gradients in that round.

The runtimes obtained for one round of the different topologies are summarized in Table \ref{mpctable}.   
\begin{table}[htbp]
\centering
\caption{Runtimes in seconds for computing secure aggregation of gradients for one round of training.}
\label{mpctable}
\begin{tabular}{c|c|c|ccc}
\hline
\multirow{2}{*}{FedAvg} & \multirow{2}{*}{DRING} & \multirow{2}{*}{DFC} & \multicolumn{3}{c}{DMS}\\
   &   &   & 3 parties  &  4 parties & 5 parties \\ \hline
 1.08s   &   0.16s   &  488.41s    &  0.16s   &  0.25s  &   0.39s \\ \hline
\end{tabular}%
\end{table}
\gls{DFC} clearly results in the most expensive secure aggregation computation, taking $488.41s\, (\approx 8\textnormal{min})$. This was to be expected, since using more \gls{MPC} parties quickly increases the communication overhead. On the other hand, for this experiment, we ran more than one party on a single machine, with some machines running up to six parties, which also contributed to the slow runtime. In FedAvg, the fact that there are only 3 \gls{MPC} parties significantly improves the performance. However, \gls{DRING} and \gls{DMS} allow even faster computation times. In DRING, each group of three neighboring agents is aggregating 3 gradients in each round (instead of the 3 \gls{MPC} parties aggregating 30 gradients in \gls{FedAvg}). If all the groups of 3 agents perform their computations in parallel, a runtime of 0.16s is achieved. Regarding \gls{DMS}, the number of agents aggregating their gradients in each round can be adjusted according to the desired runtimes. Naturally, increasing this number will also result in a slower aggregation, especially since the number of \gls{MPC} parties running the protocol will increase. Nonetheless, aggregation with 5 parties in \gls{DMS} still achieves runtimes more than twice as fast as \gls{FedAvg}.

\begin{remark}
    Due to a limitation of the SCALE-MAMBA software, each gradient to be aggregated must initially be entered as three separate values (one for the sign and two for the value itself). These values are then put together to obtain the original gradient value before performing the aggregation itself. However, this means that we need to secret share 3 times as many values, resulting in runtimes 2 to 3 times slower than if only the value itself was secretly shared. The runtimes in Table \ref{mpctable} assume that the gradients can be read as one single secret value. 
\end{remark}

\R{Due to the use of different network topologies, models and experimental setups, comparing the numerical benchmarks in Table \ref{mpctable} with those of previous related work is not straightforward. Instead, we look at the computation and communication complexities for performing secure aggregation for the communication topologies considered in this work. We compare them to corresponding complexities for the protocol in \cite{safelearn2021}, which in turn presents lower complexities than other previous works (as is shown in \cite[Table 2]{safefl2023}). 
}

\begin{table}[htbp]
\def\arraystretch{1.3}
\centering
\caption{Computation and communication complexity per training iteration of the different topologies compared with SAFELearn \cite{safelearn2021}. $m$ is the length of the model updates and $n$ is the number of clients.}
\label{mpctable_complexities}
 \resizebox{\columnwidth}{!}{%
\begin{tabular}{c|cc|cc}
\hline
\multirow{2}{*}{Approach} & \multicolumn{2}{c|}{Computation}     & \multicolumn{2}{c}{Communication}    \\ \cline{2-5} 
                          & \multicolumn{1}{c|}{Server}  & Client & \multicolumn{1}{c|}{Server} & Client \\ \hline
SAFELearn \cite{safelearn2021} & \multicolumn{1}{c|}{$\mathcal{O}(mn)$}       &   $\mathcal{O}(m)$     & \multicolumn{1}{c|}{$\mathcal{O}(mn)$}       &    $\mathcal{O}(m)$    \\ \hline
\gls{FedAvg} & \multicolumn{1}{c|}{$\mathcal{O}(mn)$}       &   $\mathcal{O}(m)$     & \multicolumn{1}{c|}{$\mathcal{O}(mn)$}       &    $\mathcal{O}(m)$    \\ \hline
DRING                     & \multicolumn{1}{c|}{-}       &   $\mathcal{O}(m)$     & \multicolumn{1}{c|}{-}       &   $\mathcal{O}(m)$     \\ \hline
DFC                       & \multicolumn{1}{c|}{-}       &   $\mathcal{O}(mn)$     & \multicolumn{1}{c|}{-}       &   $\mathcal{O}(mn)$     \\ \hline
\gls{DMS}                       & \multicolumn{1}{c|}{-}       &    $\mathcal{O}(m)$    & \multicolumn{1}{c|}{-}       &   $\mathcal{O}(m)$     \\ \hline
\end{tabular}
 }
\end{table}

\R{ The setting of the protocol in \cite{safelearn2021} is similar to the \gls{FedAvg} topology, hence the identical complexities. With the DFC topology, the absence of a distributed central server results in high communication and computation complexities for the clients, which can become unpractical as the number of clients increases. The DRING and \gls{DMS} topology allow removing the central server while maintaining low complexities on the clients' side.
Note that the number of \gls{MPC} parties also influences the complexities but is not considered in this table. This is because in \cite{safelearn2021}, \gls{FedAvg} and DRING the number of parties is small and constant throughout the protocol. In the \gls{DMS} topology, as the complexity analysis in Section \ref{sec_alg_performance}, the worst-case scenario for \gls{DMS} involves the number of interacting clients in each round equal to the total number of clients. When using DFC, the \gls{MPC} parties are the clients, and hence their impact is already accounted for.
}

\begin{figure}[htbp]
    \centering
    \includegraphics[width=0.8\columnwidth]{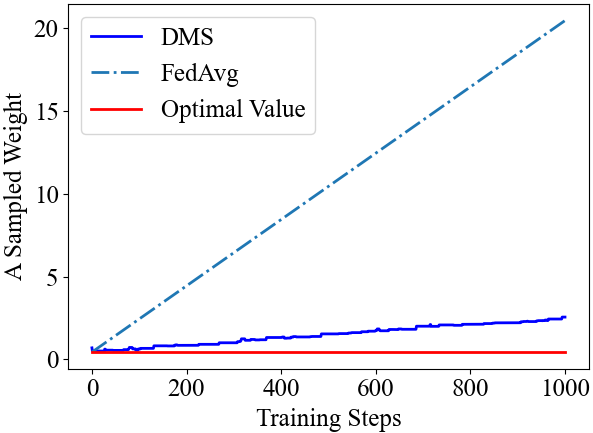}
    \caption{Weights update under attacks.}
    \label{loadfigs}
\end{figure}

\paragraph{Robustness to attacks}
\R{In this section, we conducted experiments to evaluate the robustness of our proposed method, DMS, to two different attack algorithms, poisoning attacks, and DLG. With the poisoning attack scenario, we arbitrarily and randomly attacked $10\%$ of the agents ($3$ out of $30$) and attackers introduced a poisoning attack by injecting a deliberate malicious perturbation of $0.2$ into their broadcasted model weights during each training round. We monitored the resulting mismatch between the model's parameters and the optimal value throughout the training process. The outcomes of these experiments are presented in Fig. \ref{loadfigs}, which illustrates a comprehensive comparison between our \gls{DMS} approach and FedAvg. }

\R{From the observations in Fig. 8, it is evident that our \gls{DMS} exhibits superior robustness compared to the \gls{FedAvg} algorithm under poisoning attacks. Even under a sustained onslaught of over 1000 rounds of attacks, our method displayed only a $9\%$ increase in error, unlike FedAvg. This robustness, which refers to maintaining accuracy under attack, can be attributed to our novel approach that adapts to changing communication topologies. In our implementation of the Markovian switching mechanism, only a subset of agents participate in weight updates each round, hindering attackers from consistently propagating their malicious weights. This inconsistency significantly enhances the system's robustness against adversarial influences. These results underscore the effectiveness of our \gls{DMS} method in defending against poisoning attacks, suggesting its potential to significantly enhance the security of learning systems.}



\begin{figure}
    \centering
    \includegraphics[width=0.8\columnwidth]{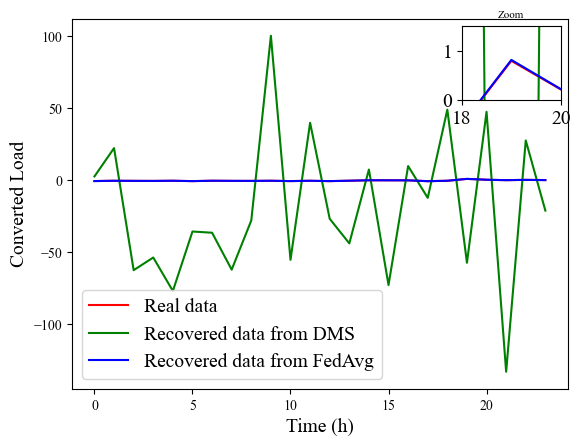}
    \caption{Recovered data from \gls{DLG} attack.}
    \label{fig:DLG_DMS_00}
\end{figure}

\R{The second scenario is under the attack of DLG. \gls{DLG} is a strong attack against gradients-sharing-based NN training. Once it hijacks the real gradients, the attacker can recover the original training set. In this experiment, an attacker randomly selects a communication line and gains continuous access to the transmitted information. By doing so, the attacker can obtain the change in weights between two communications, effectively capturing gradient information. This data is then used to recover load data. For benchmarking purposes, we compare our results against the \gls{FedAvg} algorithm in this experiment. The experiment results are shown in Fig.~\ref{fig:DLG_DMS_00}.}


\begin{figure}
    \centering
    \includegraphics[width=0.8\columnwidth]
    {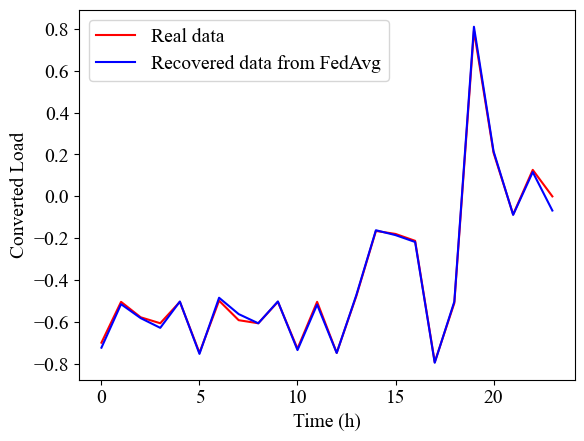}
    \caption{Comparison between recovered and real converted Load under FedAvg setting.}
    \label{fig:dlg_fl}
\end{figure}

\R{Fig.~\ref{fig:DLG_DMS_00} shows the recovered data by employing the \gls{DLG} algorithm. It can be seen that our \gls{DMS} algorithm performs better than the \gls{FedAvg} algorithm in terms of protecting the information leakage from the gradient. 
In FedAvg, the central server initializes and broadcasts the global model to clients. Therefore, each client holds an identical NN model with the same randomly initialized weights. Meanwhile, for every round of training, each client needs to share their updates with the central server. In this case, the attacker can succeed if they have access to different stages of weights, for example, the initialized weights and any weights from later updates. The comparison of recovered data under FedAvg setting can be clearly seen as~\ref{fig:dlg_fl}.
By contrast, the attacker faces more difficulties in the \gls{DMS} setting. First, the clients' models are not unified initialized. Second, the communicating clients in each round of training are randomly selected. Therefore, it is way more difficult for the attacker to hijack weights in different stages for a specific client.} 

\begin{figure}[htbp]
    \centering
    \includegraphics[width=\columnwidth]{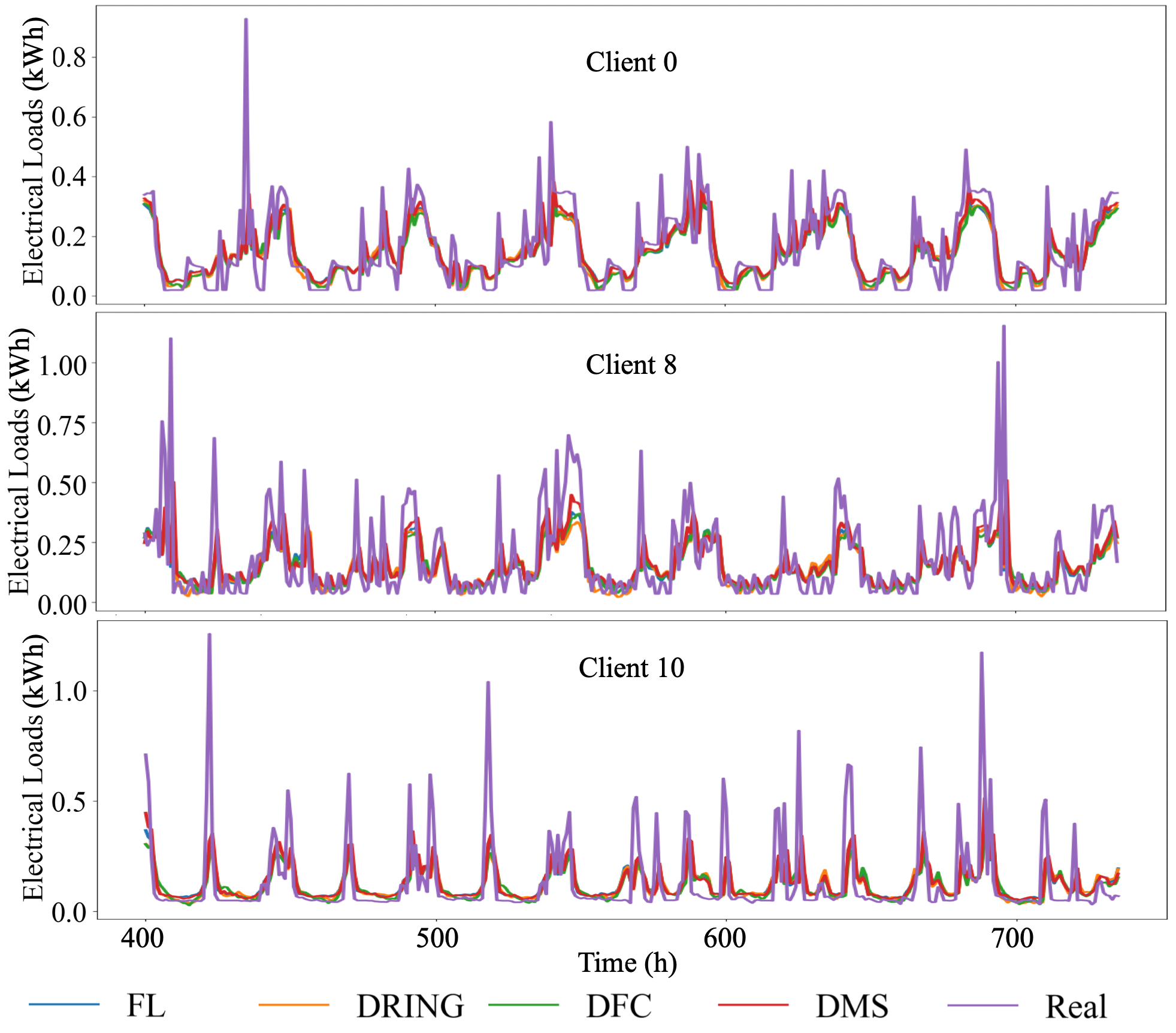}
    \caption{Forecasting performance of different algorithms on DNN models.}
    \label{loadnewfigs}
\end{figure}

\paragraph{Accuracy}
Here, we measure the accuracy of the algorithm's predictions. The evaluation metric we used is MSE. We conduct an experiment to forecast short-term residential load using four different state-of-the-art models: DNN, CNN, LSTM, and WaveNet. For each model, we implement four different algorithms: \gls{FedAvg}, \gls{DRING}, \gls{DFC} and \gls{DMS}. Firstly, the forecasting results of three randomly selected residentials are shown in Fig. \ref{loadnewfigs}, for the full forecasting results, please refer to our GitHub\footnote{\url{https://github.com/YingjieWangTony/FL-DL.git}}.

\R{From Fig. \ref{loadnewfigs}, we can see that the proposed \gls{DMS} algorithm performs well and is able to capture the trends and fluctuations in the actual electricity consumption values quite well. It is slightly better than other algorithms in most time slots. It is noticed that there are some deviations between the actual and every predicted peak value in Fig.~\ref{loadnewfigs}. That is largely due to the loss function, \gls{MSE}. Using \gls{MSE} in regression problems is common, but it has some limitations when the data is imbalanced. First, \gls{MSE} treats all errors equally, regardless if their values are high or low. In other words, \gls{MSE} will not prioritize peak value accuracy. It leads to an underestimation of peak values. Second, when the data is imbalanced, the peak values are often considered as outliers during optimization. While \gls{MSE} is sensitive to outliers, it aims to minimize the error across all samples uniformly. This can be problematic when peaks are outliers but are very important. Thirdly, global optimization could be a reason, too. The optimization process aims to minimize the global error. Peak values are rare in normal consumption profiles, the optimization process may choose to ``sacrifice" accuracy on these points in favor of better global accuracy.
We propose the following solutions to this situation, custom loss function, data resampling, feature engineering, and ensemble methods. These solutions could be a good help towards peak value prediction. However, this is beyond the current scope of the paper.}

\begin{figure}[htbp]
    \centering
    \includegraphics[width=\columnwidth]{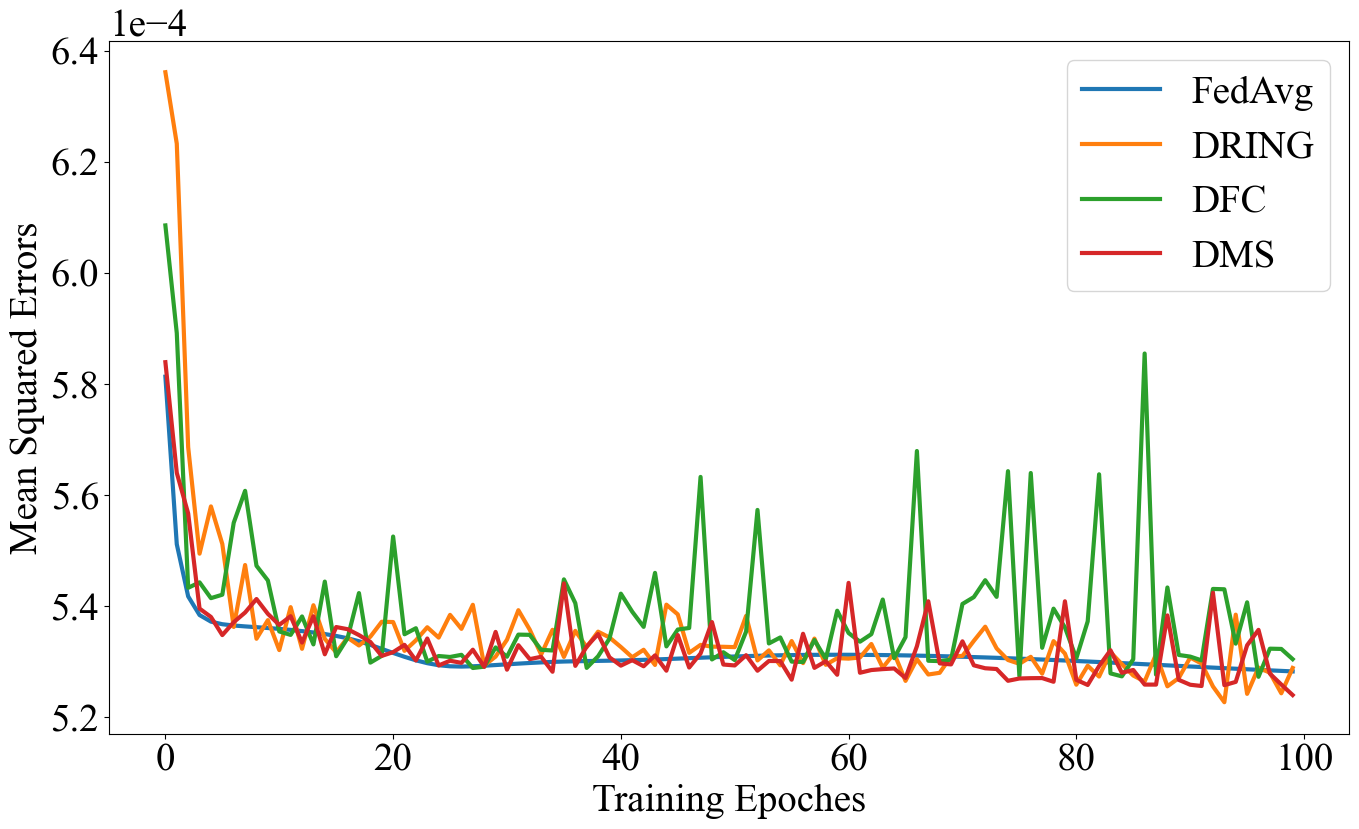}
    \caption{Training performance of different algorithms on DNN models.}
    \label{dnntraining}
\end{figure}

In Fig. \ref{dnntraining}, we show the training performance among four different algorithms. Mean squared error is applied to evaluate the performance of the models. To further illustrate the comparison results among different models and algorithms, we summarize the overall mean square error values for each model and algorithm on the testing dataset, as shown in Table.~\ref{accuracytable}.


\begin{table}[htbp]
\centering
\caption{\RR{Mean square error (kWh) of different algorithms and models.}}\label{accuracytable}
\begin{tabular}{ccccc}
\hline
         & DNN & CNN & LSTM & WAVENET \\ \hline
FedAvg        &  0.05263  &  0.06495   &    \textbf{0.0537}  &    0.4965     \\
DRING      &  0.05289   &  0.06498   & 0.0556     &    \textbf{0.3659}     \\
DFC      &  0.05305   &  0.06493   &   0.0552   &    0.5043     \\
DMS        &  \textbf{0.05236}   &  \textbf{0.06388}   &  0.0550    &   0.5156      \\ \hline
\end{tabular}%
\end{table}

From the results given in Fig. \ref{dnntraining} and Table \ref{accuracytable}, the proposed \gls{DMS} algorithm meets the requirements and has better accuracy than other models. For the dataset used in this case, the \gls{DMS} algorithm proposed obtains better performance and fewer prediction errors in most cases. 
\R{It is noticed that some fluctuations during the training process, even after 100 training epochs. These fluctuations are primarily caused by the heterogeneity in local data distributions. Each node trains on a distinct dataset, leading to models that are finely tuned to their specific data characteristics. When these models are aggregated, as in Federated Learning, the process attempts to reconcile these divergent updates, causing fluctuations in the global model. This phenomenon is exacerbated in environments where the local datasets are not independently and identically distributed. While theoretically, a centralized training approach using a powerful computing center and all available data would yield an optimal model, practical constraints like geographical limitations and privacy concerns make this approach infeasible. Thus, distributed training, despite its inherent fluctuations, becomes a necessary compromise, offering a balance between model performance and adherence to logistical and privacy constraints. Over time, as more rounds of aggregation occur, the global model tends to stabilize, gradually adapting to the diversity of local updates.}

\section{Conclusion}\label{sec_con}
In this paper, we developed a Markovian Switching distributed learning framework for residential short-term load forecasting. Moreover, a secure aggregation approach, \gls{MPC} has been employed to address the threat of deep leakage from the gradient. We analyzed \gls{DMS} from several perspectives, such as accuracy, scalability, complexity and privacy. The \gls{DMS} is compared with traditional centralized, \gls{FL}, \gls{DRING} and \gls{DFC} models. The simulation shows that the \gls{DMS} model not only secures residential-user privacy but also shows equivalent or even superior accuracy than the other models. Particularly, it significantly reduces computational complexity and enhances scalability compared to \gls{FL} models.

\bibliographystyle{IEEEtran}
\bibliography{ref}


 




\vfill

\end{document}